\documentclass[conference]{IEEEtran}
\IEEEoverridecommandlockouts
\usepackage{cite}
\usepackage{amsmath,amssymb,amsfonts}
\usepackage{algorithmic}
\usepackage{graphicx}
\usepackage{textcomp}
\usepackage{xcolor}
\usepackage{adjustbox}
\usepackage{multirow}
\usepackage{booktabs}
\usepackage{subfigure}
\def\eg{\textit{e.g}. } 
\def\ie{\textit{i.e}. }

\def\wrt{w.r.t. }
\def\etal{\textit{et al}. }

\usepackage{graphicx,multicol,subfig}
\def\BibTeX{{\rm B\kern-.05em{\sc i\kern-.025em b}\kern-.08em
    T\kern-.1667em\lower.7ex\hbox{E}\kern-.125emX}}
\begin{document}

\title{Optimizing Key-Selection for Face-based One-Time Biometrics via Morphing
{\footnotesize \textsuperscript{}}
\thanks{}
}

\author{\IEEEauthorblockN{Dail\'e Osorio-Roig$^1$, Mahdi Ghafourian$^2$, Christian Rathgeb$^1$, Ruben Vera-Rodriguez$^2$, Christoph Busch$^1$, Julian Fierrez$^2$}
\IEEEauthorblockA{1 - Biometrics and Internet Security Research Group \\ Hochschule Darmstadt, Germany \\
\{daile.osorio-roig,christian.rathgeb,christoph.busch\}@h-da.de}
\IEEEauthorblockA{2 - Biometrics and Data Pattern Analytics (BiDA) Lab \\ Universidad Autonoma de Madrid (UAM) \\ Spain \\
\{mahdi.ghafourian,ruben.vera,julian.fierrez\}@uam.es} }

\maketitle

\begin{abstract}
Nowadays, facial recognition systems are still vulnerable to adversarial attacks. These attacks vary from simple perturbations of the input image to modifying the parameters of the recognition model to impersonate an authorised subject. So-called privacy-enhancing facial recognition systems have been mostly developed to provide protection of stored biometric reference data, i.e. templates. In the literature,  privacy-enhancing facial recognition approaches have focused solely on conventional security threats at the template level, ignoring the growing concern related to adversarial attacks. Up to now, few works have provided mechanisms to protect face recognition against adversarial attacks while maintaining high security at the template level. In this paper, we propose different key selection strategies to improve the security of a competitive cancelable scheme operating at the signal level. Experimental results show that certain strategies based on signal-level key selection can lead to complete blocking of the adversarial attack based on an iterative optimization for the most secure threshold, while for the most practical threshold, the attack success chance can be decreased to approximately 5.0\%.


\end{abstract}


\section{Introduction}

Face recognition systems have been deployed in numerous access control applications, \eg border control~\cite{SmartBorders-EU-2018}, financial transactions and ID cards~\cite{William-NationalIDCards-2023}. However, the widespread use of these technologies has raised serious security and privacy concerns. Additionally, with the recent success of deep learning in facial recognition, potential adversarial attacks have been reported (\eg~\cite{Agarwal-image-agnostic-perturbations-2018,Dabouei-fast-adversarial-face-2019}). These attacks range from simple perturbation of the input image to advanced attacks in which model parameters are modified.~\cite{Xu-adversarial-face-recognition-2022}. According to Xu \etal~\cite{Xu-adversarial-face-recognition-2022}, adversarial images lead to higher false match rates when security thresholds are set in a biometric system using a clean dataset (\eg original face image without adversarial perturbation). In addition, when unauthorised subjects are allowed access to a restricted service or resource, they can launch adversarial attacks against the system and gain access to different applications~\cite{Biggio-adversarial-security-review-2015}, \eg a genuine client’s account~\cite{Biggio-security-classifiers-2013}.


 \begin{figure*}[!t]
    \centering
\includegraphics[width=\linewidth]{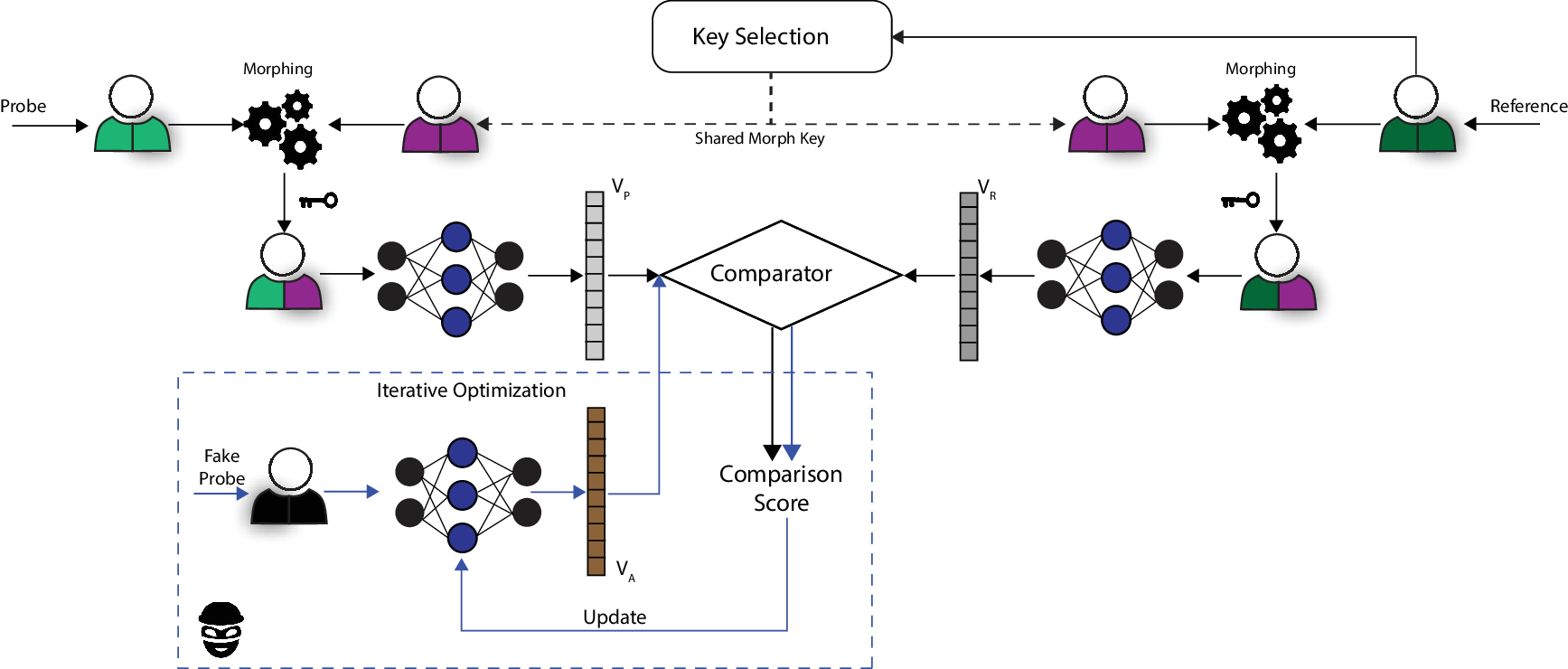}
    \caption{Conceptual overview of key selection-based OTB-morph .}   
    \label{fig:overview}
\end{figure*}

Cancelable biometrics utilise transformations in signal or feature domain which enable a biometric comparison in the transformed (encrypted) domain~\cite{Rathgeb-deep-btp-2023}, \ie biometric templates are permanently protected. In the context of cancelable biometrics, Ghafourian~\etal~\cite{Ghafourian-otb-2023} proposed a scheme that aims at protecting face templates against iterative optimization-based adversarial attacks without scarifying template protection requirements~\cite{ISO24745} such as \textit{unlinkability}, \textit{irreversibility}, \textit{renewability}, and \textit{biometric performance}. More precisely, the so-called \textit{OTB-morph} method utilises the concept of morphing attacks~\cite{Scherhag-MorphingAttacks-MorphingTechniques-BIOSIG-2017} as a transformation function for cancelable face biometrics based on time-varying keys (signal- or image-based level). The randomness employed in this transformation function (henceforth referred to as ``key'') is based on the random selection of the sample that contributes to a morph. Despite the fact that the method reduces the success chance of adversarial attacks produced by iterative optimization, it is still unknown to what extent the key selection (\ie random selection in \cite{Ghafourian-otb-2023}) in OTB-morph could lead to higher security against such attacks.


 Motivated by the above facts, this work investigates and proposes different key selection strategies for the OTB-morph algorithm ~\cite{Ghafourian-otb-2023}. In particular, we analyse how the probability of accepting the attacks produced by iterative optimization can be decreased by varying the selection strategy of a sample that contributes to a morph (\ie key selection). While OTB-morph~\cite{Ghafourian-otb-2023} utilises random sampling to produce a morphed face at the signal level, we exploit the properties of opposite demographic groups and dissimilarities of samples to generate morphed facial images. Said demographic properties lead to statistical assumptions already known in the literature~\cite{Howard-HomogeneityEffect-2019}: facial recognition algorithms produce higher similarity scores and, hence, significantly more false matches for subjects sharing similar demographic attributes, \eg gender and skin colour. Therefore, solutions that exploit the properties of opposing demographic groups are expected to contribute to a decrease in false matches. The findings of this work also lead to a better understanding of how signal-level cancelable facial biometrics  can reduce the vulnerability of biometric systems against iterative optimization-based adversarial attacks. 
 
 
 The remainder of this paper is organised as follows: Sect.~\ref{sec:related-work} briefly introduces the related work. In Sect.~\ref{sec:key-otb-morph}, a detailed explanation of OTB-morph algorithm based on different key selections is provided. Sect.~\ref{sec:metrics} presents the experimental setup and the achieved results are reported in Sect.~\ref{sec:results}. A summary of the findings is finally provided in Sect.~\ref{sec:conclusions}.

\section{Related works}
\label{sec:related-work}

This section provides a brief overview of cancelable schemes applied to biometrics. In order to improve the security and privacy of verification scenarios, the concept of cancelable biometrics was first introduced by Ratha \etal \cite{Ratha-security-privacy-2001}. In particular, a cancelable face recognition system was introduced using image warping to transform biometric data in the signal domain. Until now, many other popular cancelable techniques have been developed for multiple biometric characteristics  based on the application of non-invertible transformations, see\cite{Patel-cancelable-review-2015}. Over the past years, the majority of these transformations have been improved and, most recently, cancelable transformations have been designed to work with deep neural networks (DNNs) architectures~\cite{Rathgeb-deep-btp-2023,Shahreza-recognition-biohashing-dnn-2021}. Most of these approaches have been focused on the feature extraction step while preserving competitive biometric performance (\ie discriminatory feature space) and high privacy protection. It is worth noting that face biometrics has recently been one of the biometric characteristics that has raised the most privacy concerns. In this context of privacy, some security gaps have been analysed on cancelable face recognition systems, \eg \cite{Wang-interpretable-security-analysis-2021,Ghammam-cryptanalysis-2020}. Also, several authors have addressed these gaps by introducing hybrid protection schemes. Recently, Otroshi-Shahreza \etal \cite{Otroshi-HybridCBHE-IJCB-2022} investigated the hybrid protection by combining cancelable biometrics and homomorphic encryption.  

Ghafourin \etal \cite{Ghafourian-otb-2023} proposed a novel time-varying cancelable scheme called OTB-Morph using the morphing concept as a cancelable transformation. The authors showed full protection of cancelable deep face templates against so-called iterative optimization-based adversarial attacks.

 In summary, most of the cancelable proposals described above and existing in the literature have been analysed from different security points of view, focusing on the template level (\eg \cite{Dong-security-risks-2019}) and ensuring compliance with the requirements defined by the ISO/IEC 24745 standard~\cite{ISO24745}: \textit{unlinkability}, \textit{irreversibility}, \textit{renewability}, and \textit{biometric performance}. Moreover, recent research has studied adversarial attacks on state-of-the-art deep facial recognition systems~\cite{Xu-adversarial-face-recognition-2022,Vakhshiteh-adversarial-survey-2021}, revealing the vulnerability of facial biometric systems. To the best of the authors' knowledge and as mentioned in~\cite{Ghafourian-otb-2023}, OTB-morph has been the first work of cancelable biometric template protection scheme addressing the security threat against adversarial attacks based on iterative optimization. For comprehensive surveys on cancelable biometrics, the interested reader is referred to~\cite{Rathgeb-deep-btp-2023,Patel-cancelable-review-2015,Manisha-cancelable-comprehensive-survey-2020, Rathgeb-BTP-Survey-EURASIP-2011}.


\begin{figure*}[!t]
    \centering
  \begin{tabular}[c]{ccccc}
    \multirow{2}{*}[43pt]{
    \subfigure{\includegraphics[width=0.21\linewidth]{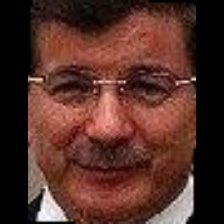}}} &
    \subfigure{\includegraphics[width=0.10\linewidth]{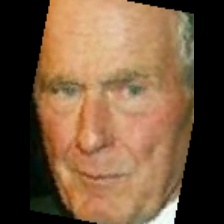}} &
    \subfigure{\includegraphics[width=0.10\linewidth]{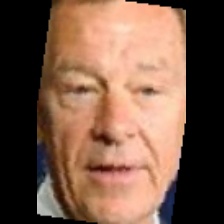}} &   
    \subfigure{\includegraphics[width=0.10\linewidth]{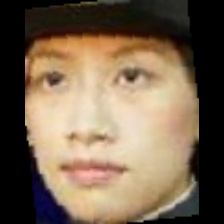}} &    
    \subfigure{\includegraphics[width=0.10\linewidth]{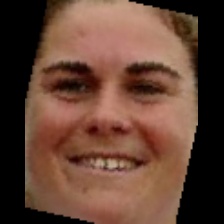}}  \\
    & \subfigure{\includegraphics[width=0.10\linewidth]{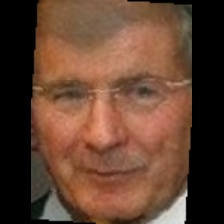}}  & 
    \subfigure{\includegraphics[width=0.10\linewidth]{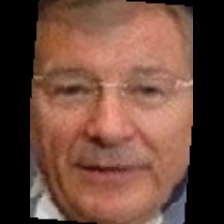}} &   
    \subfigure{\includegraphics[width=0.10\linewidth]{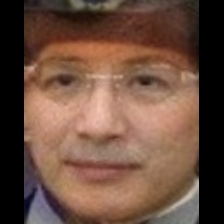}} &     
    \subfigure{\includegraphics[width=0.10\linewidth]{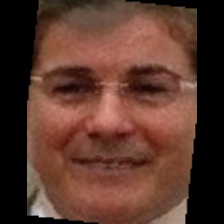}}   
     \end{tabular} 
    \caption{Examples of morph images ($2^{nd}$ row) resulting from morphing a reference image (large image on the left) with each of the samples selected by the proposed key selection strategies ($1^{st}$ row). From left to right: Random\_key, Distance\_key, SFdistance\_key, and SFrandom\_key.}
    \label{fig:key_selection}
\end{figure*}
\section{OTB-Morph}
\label{sec:key-otb-morph}
One-time-biometric via morphing (in short OTB-morph) is a new cancelable method to withstand iterative optimization attacks in face verification~\cite{Ghafourian-otb-2023}. Inspired by the ultimate security of a one-time pad~\cite{Rubin-one-time-1996} in conventional cryptography literature, this method takes advantage of morphing as a transformation function using time-varying keys (biometrics in this case) to generate protected templates at each verification attempt (Sect.~\ref{sec:key}).
Fig.~\ref{fig:overview} shows a conceptual overview of a verification scenario protected by cancelable biometrics designed at the signal level (\ie OTB-morph) that can be circumvented by an adversarial attack (\eg iterative optimization-based attack). In this attack context, the non-authorised subject has presented face images (fake probes) to the biometric system and observes the obtained comparison score and decision made by the biometric system. Finally, the iterative optimization-based attack (Sect.~\ref{sec:iterative-optimization}) is injected into the system when a biometric claim is made.

\subsection{Operation mode}
\label{sec:key}
An authentication attempt using OTB-morph is executed as follows: initially, a biometric claim is made; the facial image corresponding to the probe is morphed with another sample at the signal level using the OTB-morph approach, here, a key selection-based transformation function is employed signal-level morphing using the probe as the reference image; subsequently, the morphed facial image (morphed probe) is processed and a set of protected features is extracted using a deep neural network (DNN); subsequently, these features extracted from the morphed probe can be compared against features stored (a biometric reference) in the biometric system. Note that the features corresponding to the biometric reference have already been processed by the OTB-morph approach in an enrolment process. Also, it should be noted that the probe and the biometric reference share the same key selection for the morphing process. Finally, the biometric system verifies whether the claim is genuine (\ie the user’s identity (probe) is the one being claimed) or not (\ie the user is an impostor trying to impersonate another user), and only allows access in the former case.


In our work, facial images are morphed according to different criteria to select the sample that can contribute to a morph: $1)$ by randomly choosing a single sample (Random\_key); this type of selection has been used by OTB-Morph in the original paper~\cite{Ghafourian-otb-2023}; $2)$ by selecting the most dissimilar sample (Distance\_key); to that end, a dissimilarity score comparator is applied to the feature or embedding space to compute a distance ($s$); $3)$ by choosing the most dissimilar sample from the opposite demographic group (SFdistance\_key); for this type of criteria, the demographic information statistics (\eg gender) should be measured; $4)$ similar to criteria $3)$ but randomly selecting the sample from the opposite demographic group (SFrandom\_key). Examples of resulting images for each strategy are shown in Fig.~\ref{fig:key_selection}.  

\subsection{Iterative optimization}
\label{sec:iterative-optimization}

The idea behind this attack is to minimize a dissimilarity distance $s$ between the victim's $V_R$ face template (\ie biometric reference) and the attacker's $V_A$ face template (Fake Probe in the Fig.~\ref{fig:overview}).Therefore, for each leaked score from an impersonation attempt, the attacker updates $V_A$ with an adversarial perturbation $p_a$ such that the dissimilarity score $s$ is minimized: $\min_{s}|V_R-V_A|$. At each iteration, the objective function ($\min_{s}$) will be updated until the closest attack sample to $V_R$ is found, \ie until an impersonation attempt is successful. Note that iterative optimization-based adversarial attacks are well-known in the literature working with machine learning techniques (\eg deep learning) and can be easily optimized using the gradients of DNNs, \eg \cite{Zhu-deep-gradients-2019}.

\section{Experimental Setup}
\label{sec:experiments}

In this section, the metrics used to evaluate the different key selection strategies as well as some implementation details are summarised (Sect.~\ref{sec:metrics}). Databases and protocols employed in the assessment are also outlined (Sect.~\ref{sec:databases-protocols}).  

\subsection{Metrics and implementation details}
\label{sec:metrics}

Similar to~\cite{Ghafourian-otb-2023}, AdaFace~\cite{Kim-adaface-2022} was utilised as face feature extractor. Euclidean distance was utilised as a dissimilarity score comparator. For the morphing image process, we use the Dlib~\cite{King-dlib-2009} implementation for landmark detection and OpenCV as the morphing tool following the same settings as in~\cite{Ghafourian-otb-2023}. In particular, the morphing technique was applied directly to full-face images. The transformed facial images are then aligned and cropped using the open-source RetinaFace\footnote{https://github.com/serengil/retinaface} software. As mentioned in Sect.~\ref{sec:key}, these experiments take into account four different criteria for applying the morphing technique: random selection (henceforth referred to as Random\_key), the most dissimilar sample (henceforth referred to as Distance\_key), the most dissimilar sample from the opposite gender (\ie female or male) (henceforth referred to as SFdistance\_key), and random selection from the opposite gender (henceforth referred to as SFrandom\_key). 

The biometric performance is computed in a typical verification scenario compliant with the metrics defined in the ISO/IEC19795-1:2021~\cite{ISO-IEC-19795-1-Framework-210216} standard. The Equal Error Rate (EER), which represents the operating point at which False Match Rates (FMR) and False Non-Match Rates (FNMR) are equal, is reported. In addition, the FNMR values for several security thresholds, \ie 0 $\leq$ FMR $\leq$ 40 are depicted as Detection Error Trade-off (DET) curves. We also analysed the Attack Success Rate (ASR) which is defined by computing the number of adversarial samples that were accepted by the system at fixed security threshold. 

\subsection{Databases and protocols}
\label{sec:databases-protocols}

Experiments are conducted on well-known face datasets such as VGGFace2~\cite{Cao-vggface2-2018} and LFW~\cite{Huang-wild-2007}. The former is utilised for the biometric performance evaluation and execution of the attack. More specifically, 50 identities were selected from its test set; each identity consists of 88 samples. To evaluate biometric performance, 28 samples per identity are randomly selected, resulting in 50$\times$14 mated comparisons and 50$\times$91 non-mated comparisons. To conduct the attacks, the remaining 60 samples (30 references and 30 probes) are used per identity. In this case, the iterative optimization adversarial-based attack explained in Sect.~\ref{sec:iterative-optimization} is performed on 30 samples from the same identity. Note that the morph key generated by key selection-based OTB-morph is changed at every face verification attempt.

LFW is used for the morphing process and the application of the different key selection criteria described in Sect.~\ref{sec:metrics}. In this context, a single sample per identity with the highest quality value estimated by the CR-FIQ framework\footnote{https://github.com/fdbtrs/CR-FIQA} is selected, resulting in a total of 5,749 images.


\begin{figure}[!t]
    \centering
    \subfigure{\includegraphics[width=0.8\linewidth]{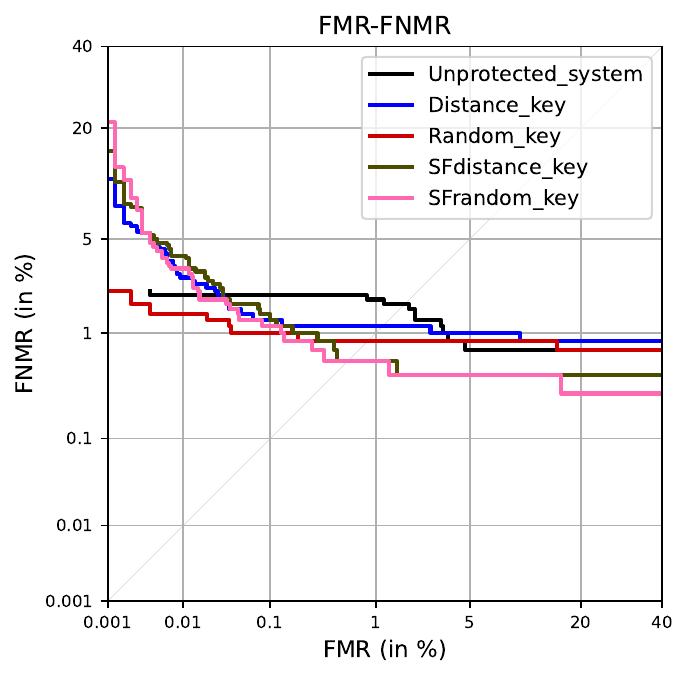}}  
    \caption{Biometric performance for different key selections.}
    \label{fig:det-curves}
\end{figure}

\begin{table}[!t]
	\scriptsize
	\caption{Error rates (in \%). The best results are highlighted in bold.}
	\label{tab:error-rates}
    \begin{adjustbox}{max width=\linewidth}
    \begin{tabular}{ccccccc}
    \toprule
        \textbf{System}&\textbf{Selection of key} & \textbf{EER}&\textbf{FMR} &\textbf{FNMR} &\textbf{Threshold}& \textbf{ASR} \\ \cmidrule{1-7}


           \multirow{4}{*}{Unprotected}&- &\multirow{4}{*}{1.71}& 0.0010 &2.00&1.1981 &1.67\\

                                     & - &                   & 0.0100 & 2.00 &1.2342 & 6.33\\
                            
                                     & - &                   & 0.1000 & 2.00 &1.2667 & 18.20\\
                            
                                     & - &                  & 1.0000 & 1.86 &1.3074 & 40.73 \\ \cmidrule{1-7}

                                    
         \multirow{17}{*}{OTB-morph}&\multirow{5}{*}{Random\_key }&\multirow{5}{*}{0.86}& 0.0010 & 2.14 &1.1543 & 0.47 \\
    
                                    &          &                      & 0.0100 & 1.42 &1.1894 &2.07 \\
                            
                                    &          &                      & 0.1000 & 1.00 &1.2292 &8.93 \\
                            
                                    &           &                     & 1.0000 & 0.86 &1.2781 & 29.60\\ \cmidrule{2-7}
                                    
                                    
                                    &\multirow{4}{*}{Distance\_key}           &\multirow{4}{*}{1.14}& 0.0010 & 11.29 &1.0751 & \textbf{0.00} \\
    
                                    &            &                             & 0.0100 &2.71 &1.1811 & \textbf{0.60}\\
                            
                                    &            &                              & 0.1000 & 1.29 & 1.2317 & \textbf{5.87}\\
                            
                                    &            &                               & 1.0000 & 1.14 &1.2818 & \textbf{25.33} \\ \cmidrule{2-7}      

                                    
                                    & \multirow{4}{*}{SFdistance\_key}          &\multirow{4}{*}{0.57}& 0.0010 & 15.57 &1.0370 & \textbf{0.00} \\
    
                                    &            &                             & 0.0100 &3.86 &1.1451 & 1.00\\
                            
                                    &            &                              & 0.1000 & 1.29 & 1.2043 &7.60 \\
                            
                                    &            &                               & 1.0000 & 0.57 &1.2566 & 25.40 \\ \cmidrule{2-7}

                                    
                                    &  \multirow{4}{*}{SFrandom\_key}          & \multirow{4}{*}{0.57}& 0.0010 & 15.57 &0.9925 & \textbf{0.00} \\
    
                                    &            &                             & 0.0100 &3.86 &1.1443 & 1.13\\
                            
                                    &            &                              & 0.1000 & 1.14 & 1.2071 & 8.87\\
                            
                                    &            &                               & 1.0000 & 0.57 &1.2567 & 28.07\\

    \bottomrule
     \end{tabular}
     \end{adjustbox}
\end{table}

\section{Results and discussion}
\label{sec:results}

Fig.~\ref{fig:det-curves} benchmarks the biometric performance of different key selection strategies as well as the unprotected system (\ie baseline). Note that all proposed key selection criteria outperform the unprotected system at the most commonly used security threshold (\ie FMR=0.1\%). For higher security thresholds (\eg FMR=0.01\%), the performance yielded by all cancelable schemes is still comparable to the unprotected system. 

Tab.~\ref{tab:error-rates} also shows the biometric performance, as well as the ASR values per key selection strategy and security threshold. Note that the key selection process assists in reducing the chances of attack compared to an unprotected system. In particular, for a threshold fixed at FMR=0.1\%, the attack chance on protected systems is approximately seven times lower than the one achieved by the unprotected system. For stricter security thresholds, the protected scheme based on Random\_key is vulnerable \wrt other key selections: Distance\_key, SFdistance\_key, and SFrandom\_key report a ASR = 0\% for a threshold fixed at FMR=0.001\%, while a slight increase above 1.0\% is observed for FMR=0.01\%. 

\begin{figure}[!t]
    \centering
    \subfigure{\includegraphics[width=\linewidth]{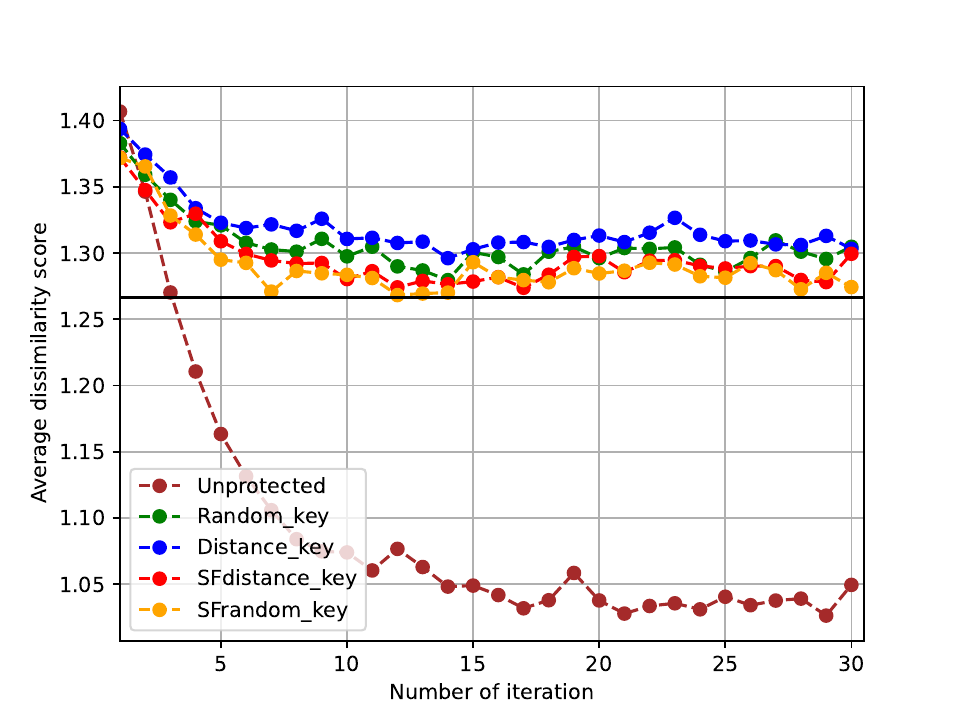}}  
    \caption{Evolution of the average comparison score achieved by the attacker. Horizontal black line visualizes the security threshold fixed at FMR=0.1\% in the baseline (\ie unprotected system).}
    \label{fig:info-loss}
\end{figure}

Fig.~\ref{fig:info-loss} reports the average dissimilarity score achieved by the attacker (\ie evolutionary process) across 30 different verification attempts. Note that the dissimilarity score computed by the unprotected system gradually decreases across the iterations, thus indicating that the attacker will be accepted by the unprotected system after a few iterations or attempts for a security threshold fixed at FMR=0.1\% (horizontal black line). Contrary to the trend shown by the unprotected system, no drastic changes are observed in the trend computed by the different key selection strategies. Note that such trends remain constant and above the security threshold in most iterations.  

\begin{figure*}[!t]
    \centering
    \subfigure[FMR=0.001\%]{\includegraphics[width=0.3\linewidth]{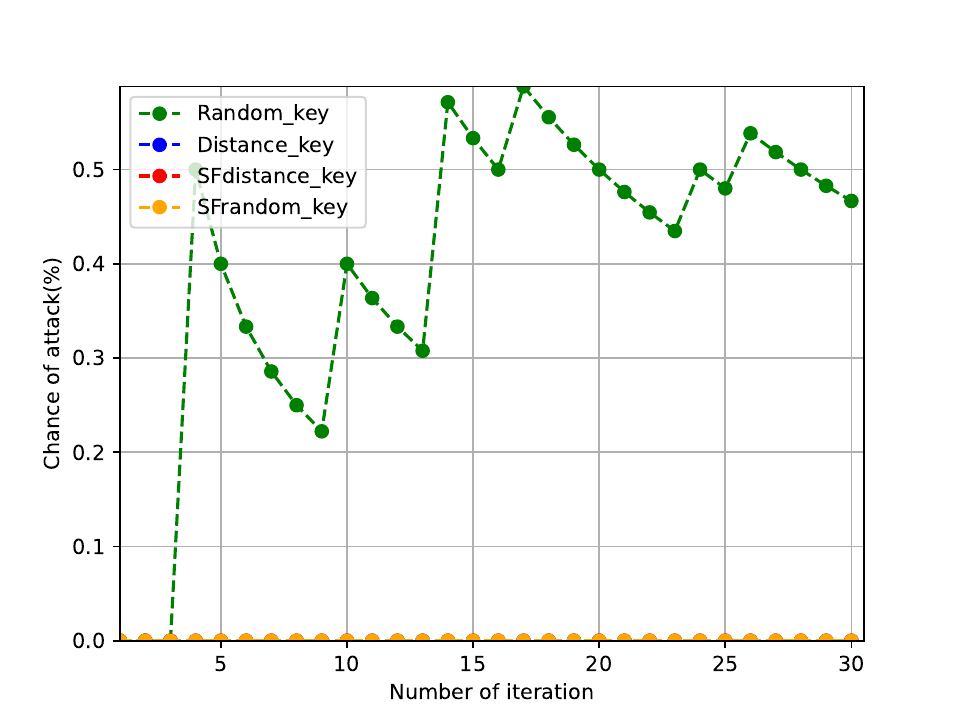}}
    \subfigure[FMR=0.01\%]{\includegraphics[width=0.3\linewidth]{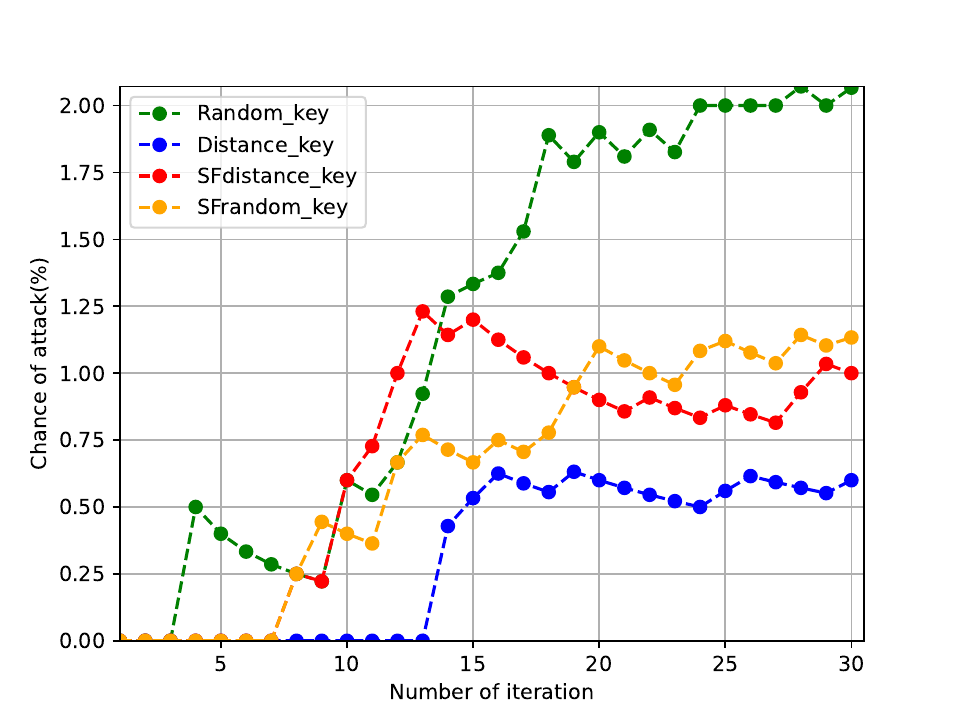}}
    \subfigure[FMR=0.1\%]{\includegraphics[width=0.3\linewidth]{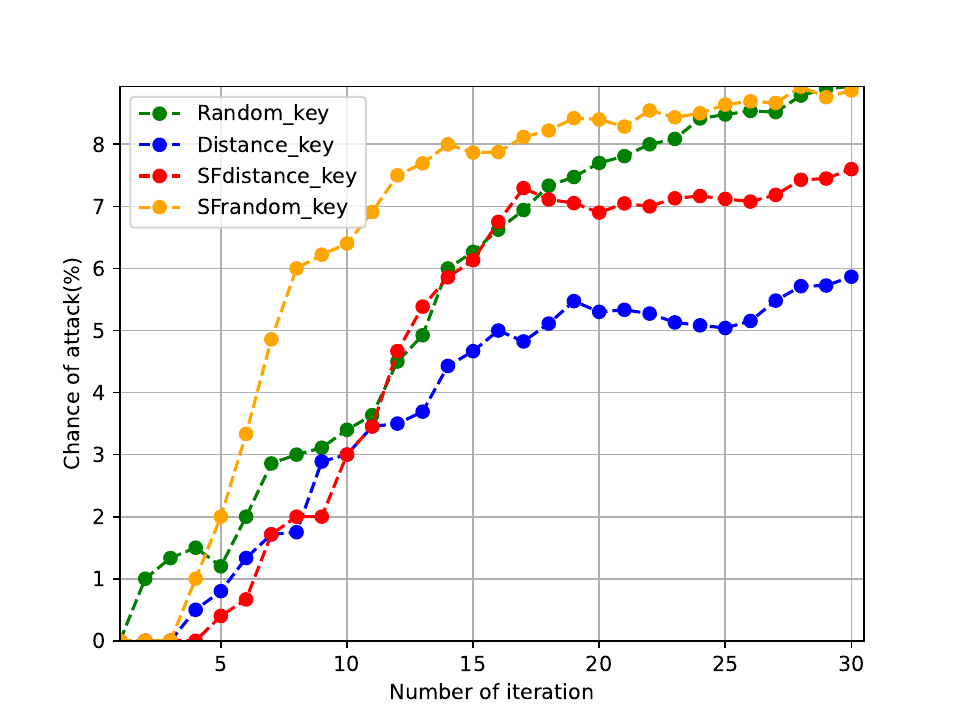}}\\    
    \subfigure[FMR=1.0\%]{\includegraphics[width=0.33\linewidth]{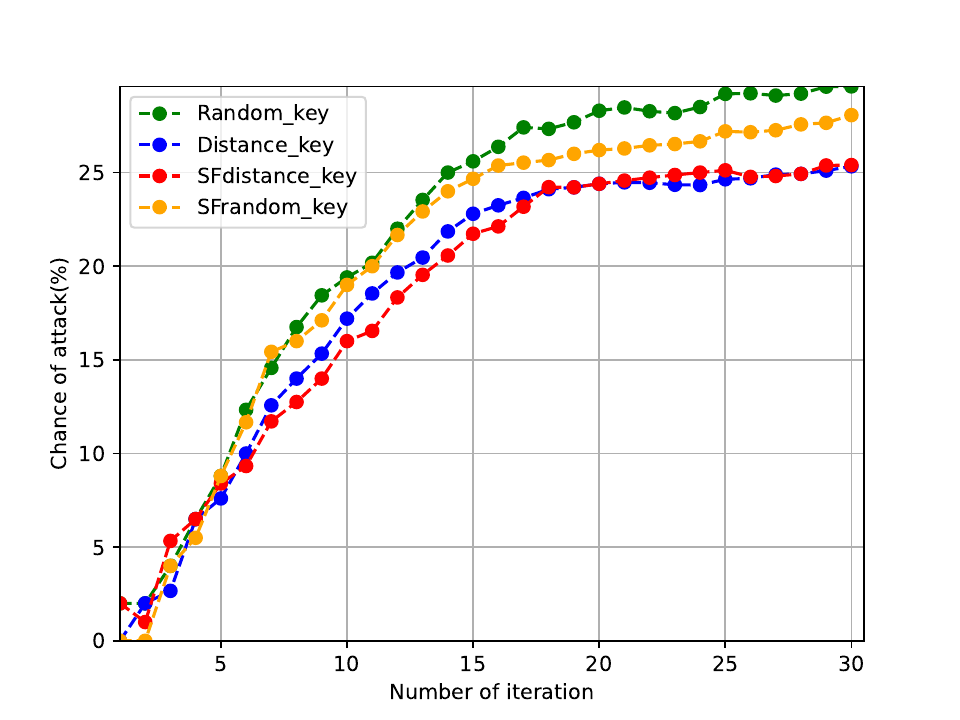}}  
    \subfigure[FMR=FNMR]{\includegraphics[width=0.33\linewidth]{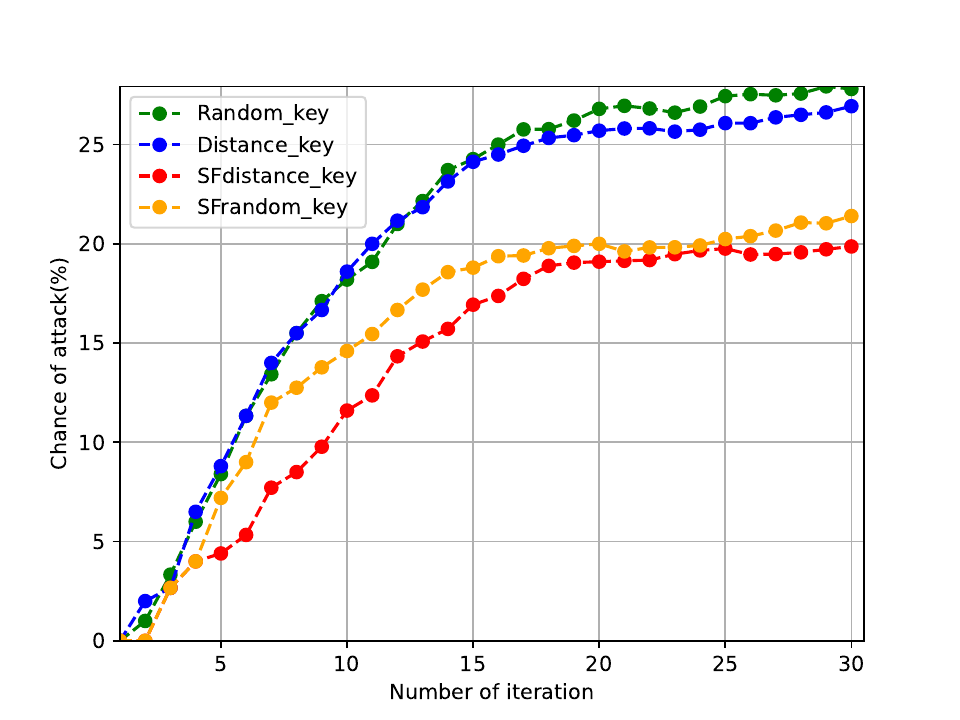}}     

    \caption{Cumulative attack chances across different numbers of iterations. Note that FMR=FNMR is defined in the threshold fixed at the EER.}
    \label{fig:accumulative-process}
\end{figure*}

Fig.~\ref{fig:accumulative-process} shows the cumulative attack chances for the different evaluated systems across 30 iterations. It can be observed that the attack rates strongly depend on the security thresholds fixed in the system. In addition, these rates confirm the results presented in Tab.~\ref{tab:error-rates}: the attack chances are reduced to 0\% for most of the key selection strategies at FMR=0.001\% with the exception of Random\_key (green line in Fig.~\ref{fig:accumulative-process}a). More importantly, the constant zero behaviour (\ie complete blocking of the attack) for all attempts can be observed, in contrast to Random\_key. For FMR=0.01\% (Fig.~\ref{fig:accumulative-process}b), the attack chance appears to be constant from a certain number of iterations for some key selections (\eg the attack is constant from iteration 15 for Distance\_key (blue line), SFdistance\_key (red line), and SFrandom\_key (yellow line)). For FMR=0.1\% (Fig.~\ref{fig:accumulative-process}c), Distance\_key (blue line) appears to be more promising, while for the more relaxed thresholds (\ie for FMR=1.0\% (Fig.~\ref{fig:accumulative-process}d) and FMR=FNMR (Fig.~\ref{fig:accumulative-process}e), the key selection based on opposite demographic information (red and yellow lines) is more challenging for the attacker. In summary, for the recommended security threshold of FMR=0.1\%, the most dissimilar image (\ie Distance\_key) is the best choice to be used as a key in the morphing process.



\section{Summary}
\label{sec:conclusions}
This work has shown that cancelable biometrics working at the signal level can be resistant to adversarial attacks. More specifically, new defence mechanisms in key selection strategies working on morphing techniques were shown to drastically reduce the chances of impostors (\eg impersonation attempts) produced by the iterative optimization-based attack. An empirical evaluation (OTB-morph in this case) showed that the randomness of signal-level cancelable schemes does not usually circumvent such attacks at their optimum. Here, the knowledge of demographic information and score distances reduced the chances of attack success down to zero percent for the highest security levels in a protected face recognition system. Future work will be focused on the impact of key selection-based OTB-morph on the variation of the dissimilarity function and face embedding extractors.


\section*{Acknowledgements}
\label{sec:acknowledgements}
This work has in part received funding from the European Union’s Horizon 2020 research and innovation program under the Marie Skłodowska-Curie grant agreement No. 860813 - TReSPAsS-ETN, PRIMA (H2020-MSCA-ITN-2019-860315), BBforTAI (PID2021-127641OB-I00 MICINN/FEDER), INTER-ACTION (PID2021-126521OB-I00 MICINN/FEDER), and the German Federal Ministry of Education and Research and the Hessen State Ministry for Higher Education, Research and the Arts within their joint support of the National Research Center for Applied Cybersecurity ATHENE.
{\small
\bibliographystyle{IEEEtran}
\bibliography{IEEEabrv}
}

\end{document}